\documentclass{article}
\usepackage{spconf,amsmath,epsfig}
\usepackage{flushend}

\title{Instance segmentation of buildings using keypoints}
\name{Qingyu Li $^{1,2}$, Lichao Mou $^{1,2}$, Yuansheng Hua $^{1,2}$, Yao Sun $^{1,2}$, Pu Jin $^{2}$, Yilei Shi $^{3}$, Xiao Xiang Zhu $^{1,2}$\thanks{This work is supported by the European Research Council (ERC) under the European Union's Horizon 2020 research and innovation programme (grant agreement no. ERC-2016-StG-714087, acronym: So2Sat, www.so2sat.eu), the Helmholtz Association under the framework of the Young Investigators Group ``SiPEO" (VH-NG-1018, www.sipeo.bgu.tum.de), and the China Scholarship Council.}}
\address{%
$^{1}$ \quad Remote Sensing Technology Institute (IMF), German Aerospace Center (DLR), Wessling, Germany\\
$^{2}$ \quad Signal Processing in Earth Observation, Technical University of Munich (TUM), Munich, Germany\\
$^{3}$ \quad Remote Sensing Technology, Technical University of Munich (TUM), Munich, Germany
}
\begin{document}
%
\maketitle
\begin{abstract}
Building segmentation is of great importance in the task of remote sensing imagery interpretation. However, the existing semantic segmentation and instance segmentation methods often lead to segmentation masks with blurred boundaries. In this paper, we propose a novel instance segmentation network for building segmentation in high-resolution remote sensing images. More specifically, we consider segmenting an individual building as detecting several keypoints. The detected keypoints are subsequently reformulated as a closed polygon, which is the semantic boundary of the building. By doing so, the sharp boundary of the building could be preserved. Experiments are conducted on selected Aerial  Imagery  for  Roof  Segmentation  (AIRS) dataset, and our method achieves better performance in both quantitative and qualitative results with comparison to the state-of-the-art methods. Our network is a bottom-up instance  segmentation  method  that could well preserve geometric details.
\end{abstract}
\section{Introduction}
\label{sec:intro}
\begin{figure}[htbp]
  \includegraphics[width=\linewidth]{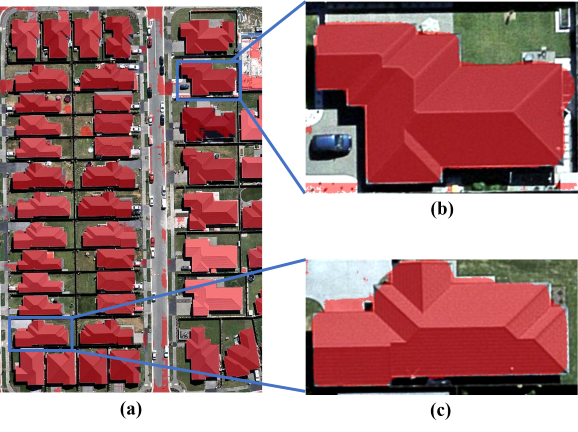}
  \caption{The building segmentation results obtained from Convolution Neural Networks (CNNs) in the large scale (a) and two zoomed in areas (b and c).}
  \label{Fig. 1}
\end{figure}

Building segmentation from remote sensing data is an important task in the remote sensing community, which benefits for a wide range of applications, such as land use management, urban planning, and monitoring. However, the variations of buildings in terms  of color, shape, material, and background bring challenges to this task. Early efforts have been made to seek out effective handcrafted visual features, for example, \cite{huang2011morphological} proposes a Morphological Building Index (MBI) to model a relation between building characteristics (e.g., brightness, size, and contrast) and morphological operators. However, these methods have poor generalization abilities. Recently, Convolutional Neural Networks (CNNs) have been widely used for building segmentation and shown promising results \cite{shi2018building} also in large-scale tasks (see Fig. \ref{Fig. 1} (a)). However, when we zoom in these segmentation results, we can clearly see that such results are not perfect, e.g., (Fig. \ref{Fig. 1} (b) and Fig. \ref{Fig. 1} (c)), where the boundary of the individual building is blurred. This is caused by the pooling layers in many existing methods which directly learn semantic masks. The presence of pooling layers have resulted in information loss, which further reduces the chance of preserving the sharp boundary.

We have observed that buildings, which are man-made objects, usually have distinct corner points. The corner points can effectively depict the shape and structure of buildings. Therefore, in this paper, we proposed a bottom-up
instance segmentation method that firstly detects keypoints of a building, and then reconstructs semantic masks with these keypoints. By doing so, more fine-grained boundaries of buildings could be preserved. 

We notice a contemporary work, PolyMapper \cite{li2019topological}, that also uses keypoints for building segmentation. PolyMapper predicts keypoints and groups them with a CNN-Recurrent Neural Network (RNN) structure. Our approach differs in two key aspects: keypoint detection and grouping. In PolyMapper, a heatmap mask of building boundaries are generated firstly, and then a mask of candidate keypoints is obtained from the additional convolutional layer. Our proposed method avoids this intermediate learning and directly detect keypoints from the input. Another difference to PolyMapper is the grouping. Our grouping approach is fully geometric-based grouping without any deep feature learning. 
\section{Methodology}
\label{sec:method}
\subsection{Overview}
\begin{figure*}[h]
\begin{center}
  \includegraphics[width=0.8\textwidth]{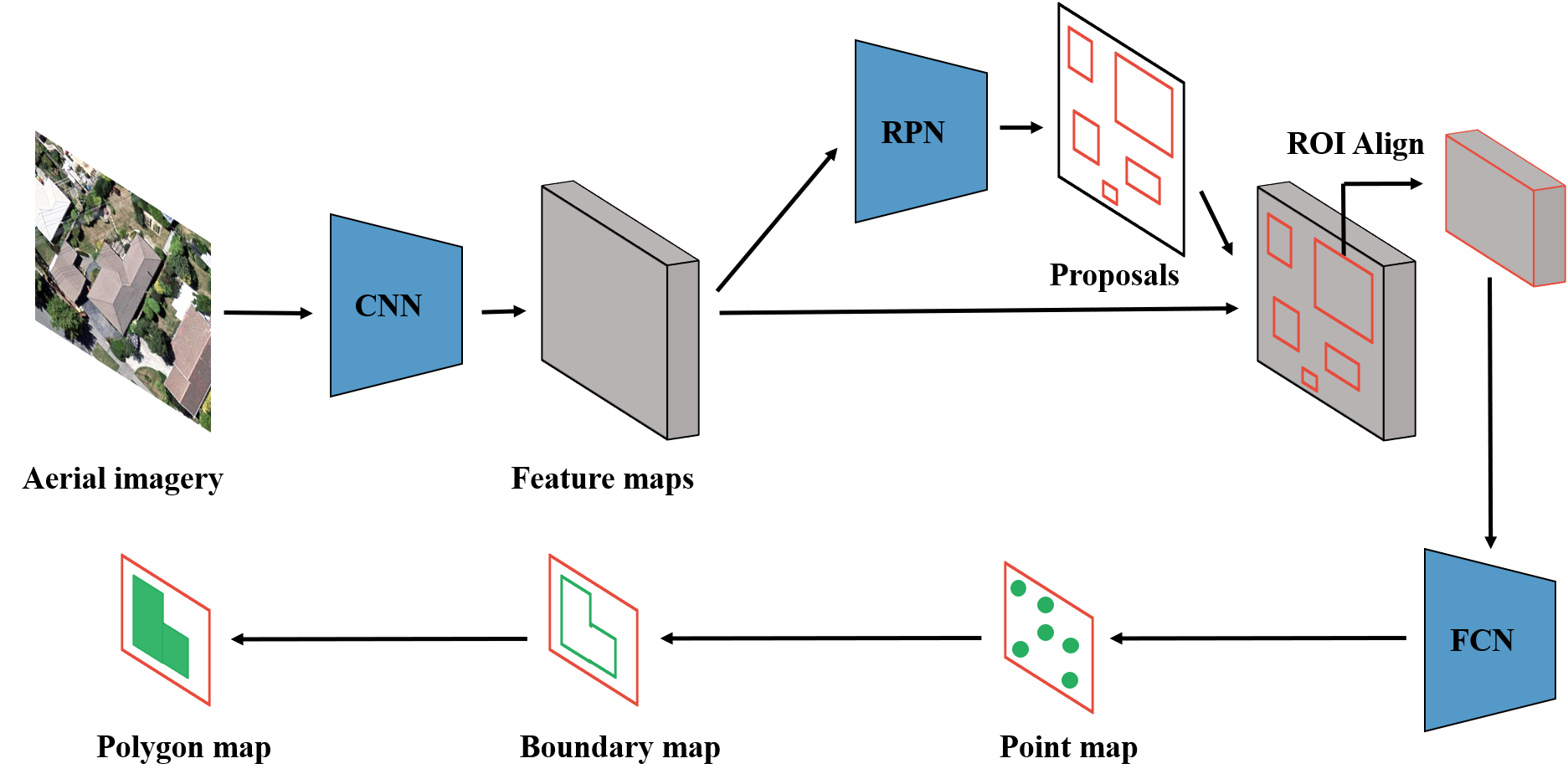}
\end{center}
  \caption{The flowchart of the proposed approach for building segmentation. CNN: Convolutional Neural Network. RPN: Region Proposal Network. FCN: Fully Convolutional Network.}
  \label{Fig. 2}
\end{figure*}

In our approach, a building is considered as a set of keypoints. Fig. \ref{Fig. 2} provides an overview of the proposed approach, which consists of a CNN, a Region Proposal Network (RPN), and a Fully Convolutional Network (FCN). Firstly, the CNN is utilized to extract feature maps, and then followed by the RPN, which slides over feature maps in order to generates “proposals” (candidate bounding boxes) where buildings may exist. For each proposal, local features are acquired by RoIAlign \cite{he2017mask}. Then FCN is applied to the features, and predicts the heatmap of the keypoints. Once the keypoints are extracted from the heatmap, they are grouped into boundaries in a purely geometric way. Finally, the buildings of interest could be delineated with these boundaries as a polygon map.
\subsection{Keypoints Detection and Grouping}
Our approach is a two-stage procedure. In the first stage, class and box offsets of proposals are predicted in parallel. Then the second stage outputs a heatmap of keypoints for each object. For each input patch, a corresponding heatmap $Y \in (0,1) ^{H \times W}$ is predicted from the proposed network, where $H$ and $W$ are the height and width of the input patch, respectively. This heatmap indicates locations of the keypoints. The training is guided by a regression of a Gaussian heatmap $P \in (0,1) ^{H \times W}$, where each keypoint denotes the mean of a Gaussian kernel. In this regard, the penalty is reduced to negative locations within a radius of the positive location instead of equal penalization during training. This is due to the fact that some false keypoint detections can still generate a bounding-box which is sufficiently overlapped with the ground reference annotation. For keypoint estimation on each object, a modified focal loss \cite{law2018cornernet} is utilized for the training, which can maintain a balance between the positive and negative locations:
\begin{equation}
\begin{aligned}
&L_{polygon}= \\
&-\frac{1}{N}\sum_{i=1}^H\sum_{j=1}^W\left\{ \begin{array}{ll}
(1-P_{ij})^\alpha\log(P_{ij}) & \textrm{if $Y_{ij}=1$} \\
(1-Y_{ij})^\beta(P_{ij})^\alpha\log(1-P_{ij}) & \textrm{otherwise} \\
\end{array} \right.
\end{aligned}
\end{equation}
, where $N$ is the number of objects in a patch, and $\alpha$ and $\beta$ are the hyper-parameters and fixed to $\alpha= 2$ and $\beta= 4$ during training. The total loss of our network is a multi-task loss $L=L_{cls}+L_{box}+L_{polygon}$. $L_{cls}$ is a cross entropy loss for bounding-box classification and $L_{box}$ is a bounding-box loss for bounding-box regression, which are defined in \cite{girshick2015fast}. 

The corresponding keypoints are extracted from the predicted heatmap by detecting all peaks. This procedure is called ExtractPeak \cite{zhou2019bottom}, where the pixel locations with a value greater than a threshold $\tau$ are firstly selected, and then peaks are the locally maximum in a window with size $3 \times 3$ surrounding these selected pixels. Here, we set $\tau=0.1$.

Finally, we adopt a simple geometric method to approximate the segmentation mask by creating a polygon where edges are sequentially connected with keypoints. More specifically, the extreme keypoint (the most left or right or bottom or top one) is firstly selected as the start point, and then the first edge would be generated by establishing a connection between it and its nearest neighbour. Then the latter is considered as start point for the next round. Edges are extended until the end keypoint meets the initial keypoint. Finally, a polygon is formed by utilizing all these generated edges.
\section{Experiment}
\label{sec:experiment}
The Aerial Imagery for Roof Segmentation (AIRS) dataset \cite{chen2019aerial} is a publicly available dataset, which aims at developing methods of building segmentation from very-high-resolution aerial imagery (0.075 meter). Note that our goal in this work is to accurately segment individual buildings. Therefore, 1680 patches each containing an individual building in the center, are extracted from AIRS dataset to validate our method. The training/validation/test split is as follows: 1400 patches for training, 140 patches for validation, and 140 images for testing, where each patch is with the size of $512 \times 512$. In this research, the proposed approach is implemented within a keras framework on an NVIDIA Tesla P100 with 16 GB of memory. The training strategy follows \cite{ohleyer2018building}, which uses SGD optimizer with learning momentum as 0.9 for 40 epochs training.
\section{Results}
\label{sec:results}
In order to evaluate the performance of our proposed algorithm, four metrics are selected in this research. The mask accuracy is evaluated by F1-Score and Intersection Over Union (IoU), while Structural Similarity Index (SSIM) and F-Measure are utilized as accuracy measures for boundary. 

\begin{figure}[htbp!]
  \includegraphics[width=\linewidth]{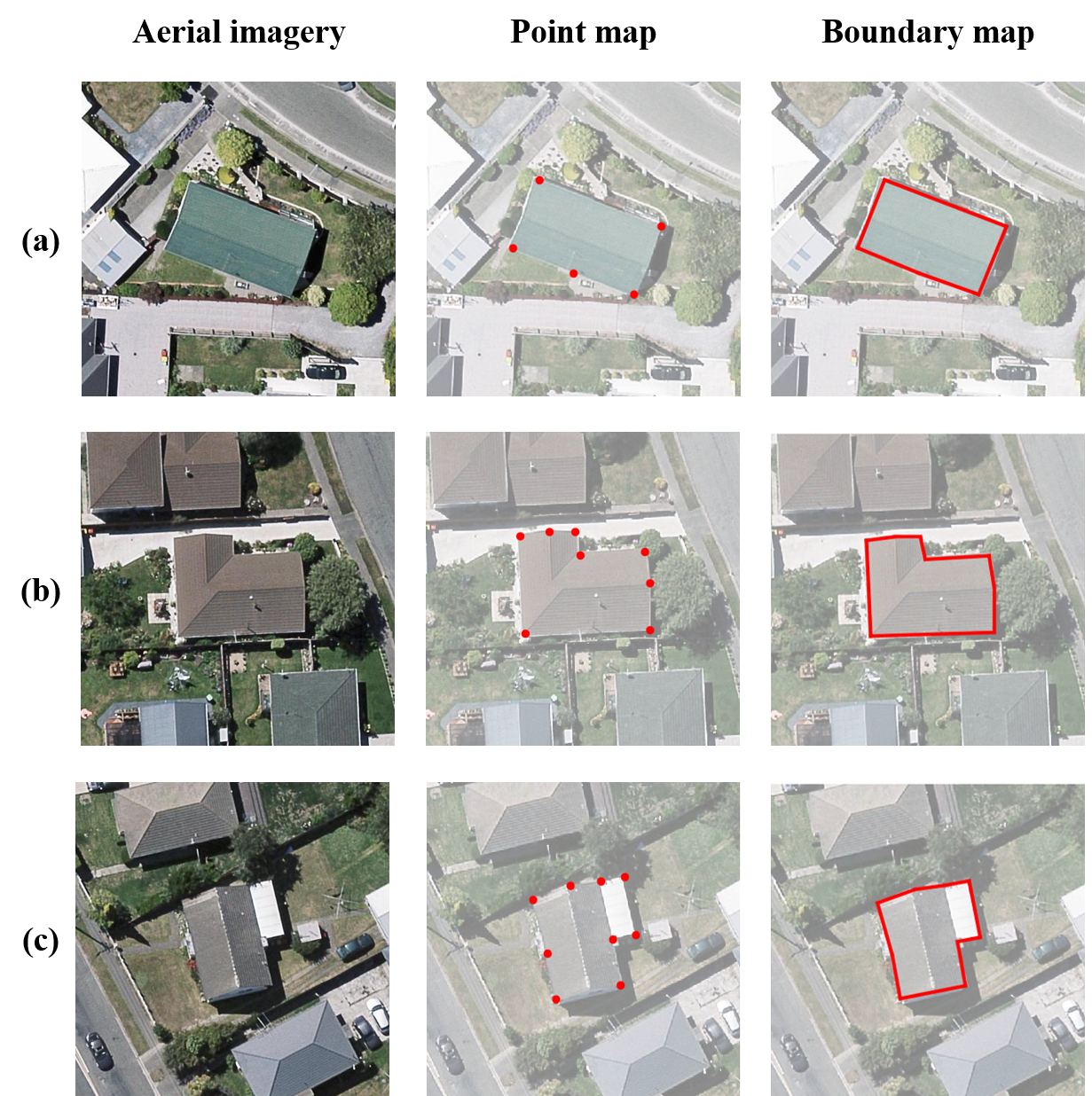}
  \caption{Aerial imagery, point map, and boundary map of the proposed method from example (a), (b), and (c).}
  \label{Fig. p1a}
\end{figure}

Fig \ref{Fig. p1a} shows the intermediate results obtained from our approach, which are overlaid on the input aerial imagery. One is the point map, which is acquired from the predicted heatmap of keypoints using the deep network. The other is the boundary map formed by these detected keypoints. 

\begin{figure*}[h]
\begin{center}
  \includegraphics[width=0.8\textwidth]{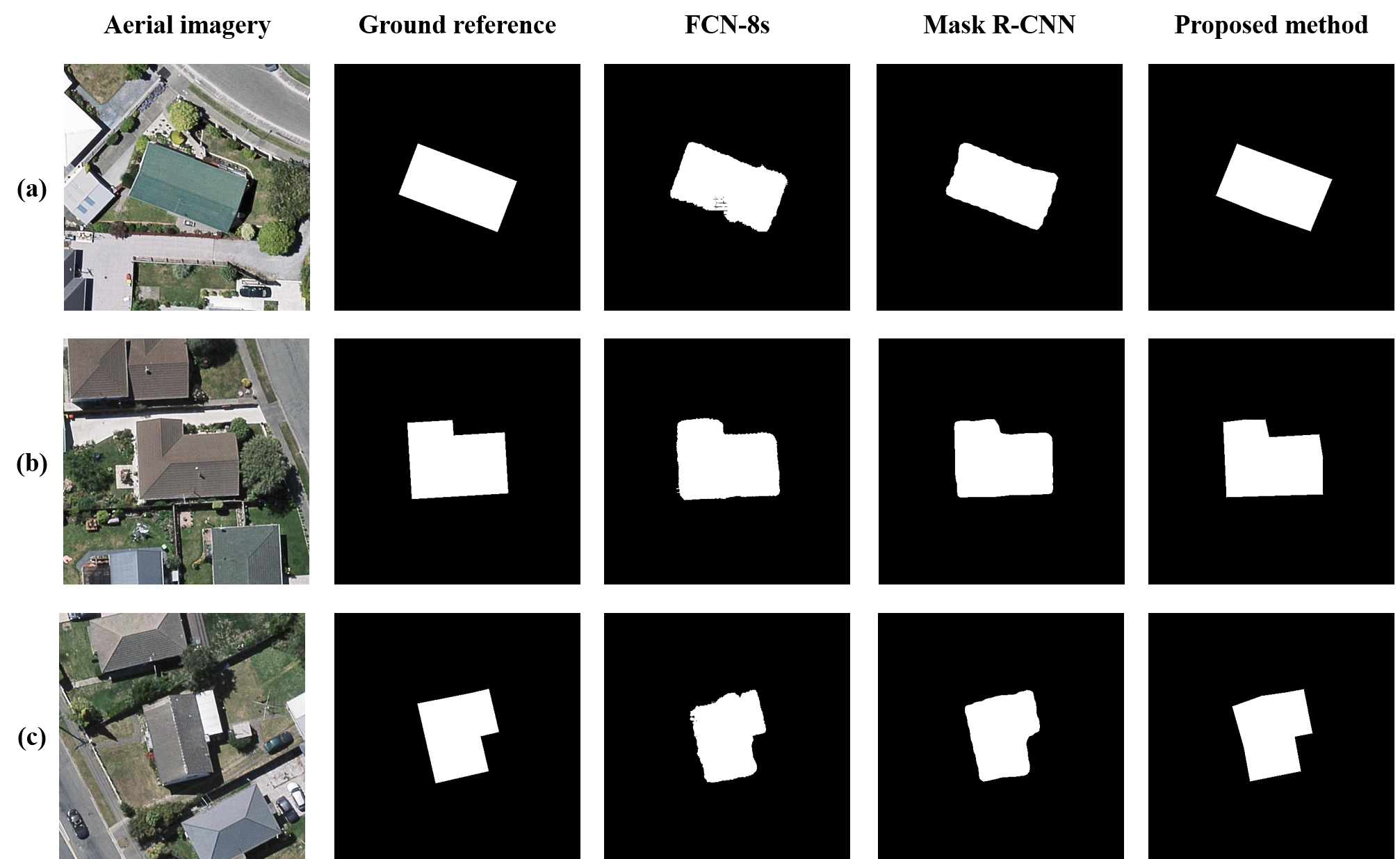}
 \end{center}
  \caption{Aerial imagery, ground reference, and visual results of different building segmentation methods from example (a), (b), and (c).}
  \label{Fig. p2a}
\end{figure*}

Table \ref{Tab. s2} and Fig \ref{Fig. p2a} compare our method to FCN-8s and Mask R-CNN, which are the state-of-the-art semantic segmentation and instance segmentation methods, respectively. The proposed approach outperforms FCN-8s and Mask R-CNN in terms of both mask and boundary accuracy. Notable, sharper boundaries and geometric details are preserved in our network, thus, the shape and structure of the buildings are well depicted. This shows the advantage of detecting keypoints over directly learning semantic masks for the task of building segmentation. 

\begin{table}[htbp!]
 \caption{Accuracy indices of different building segmentation methods}
 \begin{center}
 \setlength\tabcolsep{2pt}
 \begin{tabular}{lllll}
   \hline\hline
     ~ &\multicolumn{2}{c}{Mask} & \multicolumn{2}{c}{Boundary}\\
     \hline\hline
     Method &    F1-Score   &   IoU  &    SSIM   &   F-Measure\\
    \hline\hline
     FCN-8s & 89.01 \% & 83.15 \% & 92.58 \% & 8.29 \% \\
     Mask R-CNN & 94.73 \% & 90.22 \% & 96.82 \% & 9.63 \%\\
     \begin{bfseries}Proposed method\end{bfseries} & \begin{bfseries}95.08 \%\end{bfseries} & \begin{bfseries}90.81 \% \end{bfseries}& \begin{bfseries}96.93 \%\end{bfseries}& \begin{bfseries}11.29 \%\end{bfseries}\\
      \hline
 \end{tabular}
 \end{center}
 \label{Tab. s2}
 \end{table}
 
\section{Conclusion}
In this paper, we have proposed a new instance segmentation approach that obtains semantic mask of buildings based on keypoint detection. Our approach firstly detects keypoints of a building and then polygonize them to generate a segmentation mask with fine semantic boundaries. We evaluate our method on a selected AIRS dataset, and experimental results demonstrate that the proposed network is capable of providing competitive results compared to the state-of-the-art semantic segmentation and instance segmentation methods. It should be noted that the generated building boundaries by our method are fine grained, and shapes of buildings are well preserved in segmenation masks. This would be beneficial to further steps such as vectorization, which rely much on accurate geometric details .
\bibliographystyle{IEEEbib}
\bibliography{refs}
\footnotesize
\end{document}